\documentclass{article}
\pdfoutput=1

\usepackage[preprint,nonatbib]{neurips_2022}

\usepackage[utf8]{inputenc} 
\usepackage[T1]{fontenc}    
\usepackage{hyperref}       
\usepackage{url}            
\usepackage{booktabs}       
\usepackage{amsfonts}       
\usepackage{nicefrac}       
\usepackage{microtype}      
\usepackage{xcolor}         
\usepackage{subcaption}
\usepackage{graphicx}
\usepackage{wrapfig}
\usepackage{amsmath,amssymb}
\usepackage{algorithm, algorithmic}

\usepackage{listings}
\usepackage{xcolor}

\definecolor{codegreen}{rgb}{0,0.6,0}
\definecolor{codegray}{rgb}{0.5,0.5,0.5}
\definecolor{codepurple}{rgb}{0.58,0,0.82}
\definecolor{backcolour}{rgb}{0.95,0.95,0.92}

\lstdefinestyle{mystyle}{
    backgroundcolor=\color{backcolour},   
    commentstyle=\color{codegreen},
    keywordstyle=\color{magenta},
    numberstyle=\tiny\color{codegray},
    stringstyle=\color{codepurple},
    basicstyle=\ttfamily\footnotesize,
    breakatwhitespace=false,         
    breaklines=true,                 
    captionpos=b,                    
    keepspaces=true,                 
    numbers=none,                    
    numbersep=5pt,                  
    showspaces=false,                
    showstringspaces=false,
    showtabs=false,                  
    tabsize=2
}
\title{Hyper-Learning for \\ Gradient-Based Batch Size Adaptation}

\author{%
  Calum R. MacLellan \\
  University of Strathclyde\\
  Glasgow, United Kingdom \\
  \texttt{{calum.maclellan@strath.ac.uk}} \\
  \And
  Feng Dong \\
  University of Strathclyde\\
  Glasgow, United Kingdom \\
  \texttt{{feng.dong@strath.ac.uk}} \\
}

\begin{document}

\maketitle

\begin{abstract}
Scheduling the batch size to increase is an effective strategy to control gradient noise when training deep neural networks.
Current approaches implement scheduling heuristics that neglect structure within the optimization procedure, limiting their flexibility to the training dynamics and capacity to discern the impact of their adaptations on generalization.
We introduce Arbiter as a new hyperparameter optimization algorithm to perform batch size adaptations for learnable scheduling heuristics using gradients from a meta-objective function, which overcomes previous heuristic constraints by enforcing a novel learning process called \textit{hyper-learning}.
With hyper-learning, Arbiter formulates a neural network agent to generate optimal batch size samples for an inner deep network by learning an adaptive heuristic through observing concomitant responses over $T$ inner descent steps.
Arbiter avoids unrolled optimization, and does not require hypernetworks to facilitate gradients, making it reasonably cheap, simple to implement, and versatile to different tasks.
We demonstrate Arbiter's effectiveness in several illustrative experiments: to act as a stand-alone batch size scheduler; to complement fixed batch size schedules with greater flexibility; and to promote variance reduction during stochastic meta-optimization of the learning rate.

\end{abstract}

\section{Introduction}
Stochastic gradient descent (SGD) has become the standard algorithmic framework for training deep neural networks (DNNs) due to its computational efficiency, but inherits unique hyperparameters that affect its ability to guide DNN parameters toward good local minima. In particular, 
the batch size hyperparameter directly influences parameter movements and the properties of the minima by controlling gradient noise during optimization \cite{three-factors,bayes-sgd,how-sgd}.
Determining an optimal batch size is necessary for generalization, but remains a challenging problem within deep learning.
\\\\
Recent work has targeted scheduling heuristics for increasing the batch size from an initially small value as a function of training epochs \cite{smith-bs,adabatch,cbs}. 
However, such heuristics perform adaptations without considering information related to the batch size, thereby 
sustaining rigid characteristics that are incongruous to the stochastic nature of DNN training.
This has been effectively addressed in other works \cite{gauss,abs}, which consider the local geometry of the objective landscape when performing batch size adaptations.
Despite this, there remains a need to access deeper structural information relating the batch size to the optimization procedure of the DNN, a technique that has been used to adapt other hyperparameters such as the learning rate \cite{hd}.
With this information, the batch size could be adapted as a function of the local dynamics, and its concomitant effects on generalization, creating heuristics with significantly greater capacity to determine optimal batch sizes throughout the course of training.
\\\\
A common technique for adapting hyperparameters in this manner is through automated hyperparameter optimization (HO).
Two key HO frameworks
have emerged: (1) gradient-based, and (2) black-box approaches.
\textit{Gradient-based} approaches have direct access to the DNN optimization procedure, which enables them to tune hyperparameters from parameter responses \cite{maclaurin,hd,fran18}, but strictly demand hyperparameters to be differentiable to the objective \cite{stn}.
This is intractable for the batch size, which admits discrete and non-differentiable structure.
Furthermore, gradients are obtained from unrolled computation graphs, 
which often incur exploding gradients and truncation biases \cite{pes,short-horizon}. 
\textit{Black-box} approaches such as Bayesian optimization \cite{snoek} 
and reinforcement learning (RL) \cite{zoph} relax differentiability requirements, and learn scheduling policies for hyperparameters through a trial-and-error based observational strategy. 
However, such approaches ignore structure within the objective, resulting in the need to restart the DNN training process multiple times to form robust policies
\cite{stn}.
A black-box approach to batch size scheduling is thus theoretically possible, 
but would be invariably inefficient in practice, and unable to move in-step with the dynamics throughout training.
\begin{wrapfigure}{r}{0.5\textwidth}
\centering
    \includegraphics[width=0.5\textwidth]{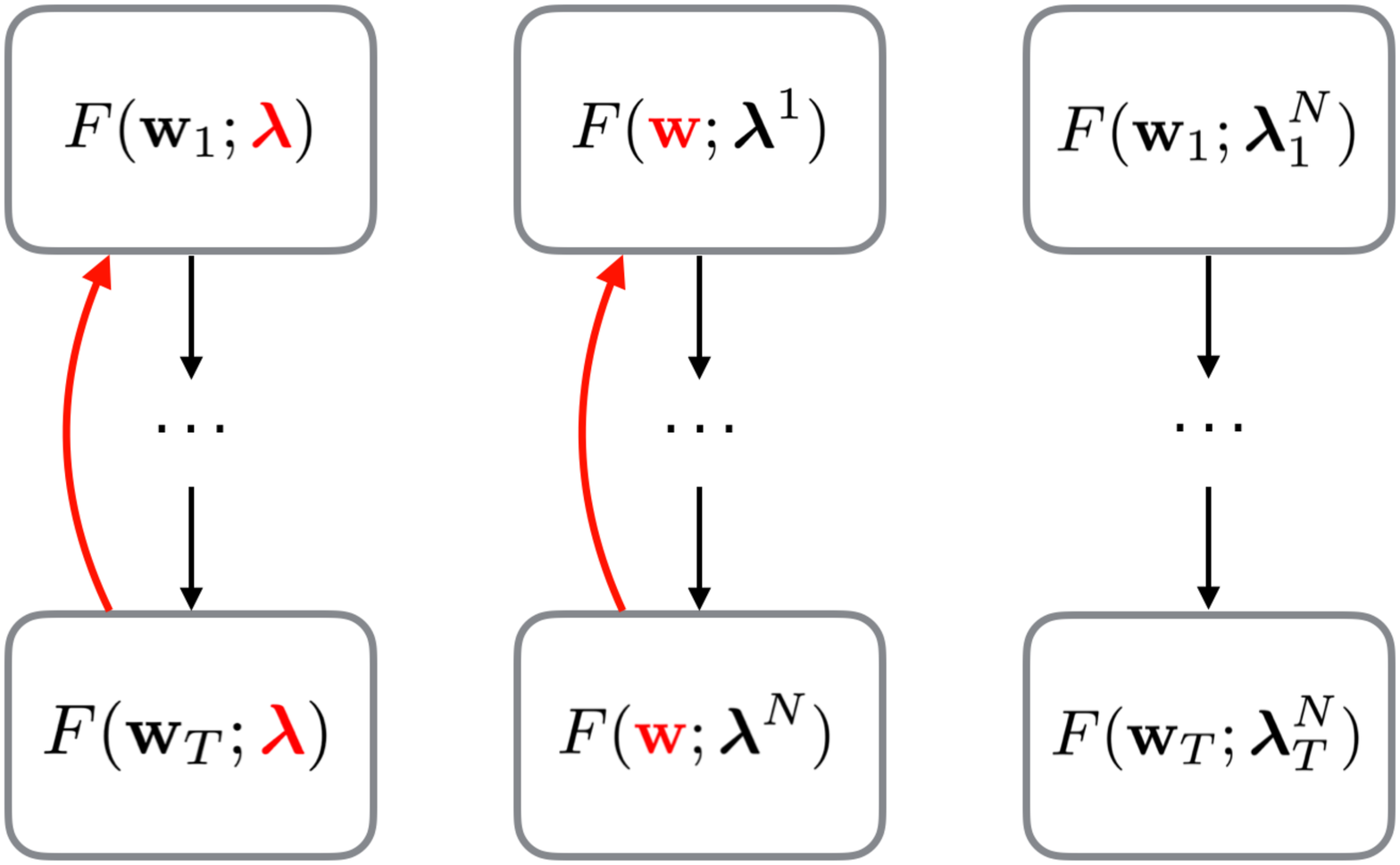}
\caption{\textbf{Hyperparameter optimization.} 
(\textbf{left}) \textit{Gradient-based} approaches perform $T$ descent steps to the weights ($\mathbf{w}$) on the meta-objective ($F$) before updating the \textcolor{red}{fixed} hyperparameter ($\boldsymbol{\lambda}$).
(\textbf{middle}) \textit{Black-box} approaches evaluate $N$ hyperparameter samples before updating the \textcolor{red}{fixed} weights.
(\textbf{right}) \textit{Hyper-learning} efficiently searches over $N$ hyperparameter samples in parallel with the $T$ descent steps to the weights.
}
\label{fig:hyperlearning}
\end{wrapfigure}
\\\\
In this paper, we design a more representative system by 
introducing a new learning process for HO called \textit{hyper-learning} (short for hyperparameter learning), which blends the data efficiency of gradients with the observational tactics of black-box methods. 
Hyper-learning formulates a neural network \textquoteleft agent'
to learn about hyperparameters by generating differentiable samples for them, and observing how they affect the dynamics of an inner \textquoteleft system' (i.e. DNN) over a period of time. 
By imposing a bilevel program between the agent and system we leverage structure from gradients, whilst avoiding unrolled optimization and training restarts by updating hyperparameters on the basis of information \textit{learned} by the agent.
Hyper-learning therefore unifies the strengths of both gradient and black-box approaches to create a structured and efficient system for searching over hyperparameter spaces. We illustrate the hyper-learning process in Figure \textbf{\ref{fig:hyperlearning}.
}

%
%
With hyper-learning, we develop an algorithm to perform gradient-based batch size scheduling, called Arbiter.
Under this framework, Arbiter's goal is to determine an optimal batch size after $T$ descent steps by predicting batch size samples at each step, and moving the samples toward optimality by observing concomitant meta-objective responses from the inner DNN. Arbiter benefits from a simple plug-and-play architecture, and because it parameterizes the batch size within the high-level features of the inner DNN, we do not require hypernetworks \cite{ha} for sustaining gradients, making it computationally cheap and scalable to deep architectures. 

\textbf{Contributions.}
\begin{itemize}
	\item We introduce a novel hyperparameter optimization (HO) algorithm called Arbiter to learn scheduling heuristics for batch size adaptations using gradients from a meta-objective. To the best of our knowledge, Arbiter is the first gradient-based HO framework developed for tuning the batch size.	
	\item We demonstrate the empirical benefits of Arbiter in various tasks: (1) to act as a stand-alone batch size scheduler; (2) to introduce greater flexibility into fixed batch size schedules; and (3) to impart variance reduction to stochastic meta-optimization of the learning rate. 
    \item Arbiter is developed under a new HO process called \textit{hyper-learning}, which formulates a neural network to learn a mapping to optimal hyperparameter samples for an inner network. Hyper-learning inherits strengths from gradient-based and black-box HO approaches, but avoids the need for unrolled optimization or training restarts.
\end{itemize}
%
\section{Related work}
\label{sec:relatedwork}

\textbf{Batch size scheduling.} Previous work has explored a range of schedules for the batch size using fixed \cite{adabatch,smith-sgd,elastic,dynamic-sgd}, dynamic \cite{ones}, and policy-based heuristics with RL \cite{rl-bs}. Improved training speed and GPU efficiency are the prevailing goals of such work, and although compelling and practical, 
are unable to provide robust answers for when and how the batch size should be adapted relative to the learning dynamics.
More representative heuristics have addressed these questions by basing adaptations on gradient variance using inner-product tests \cite{lbfgs,adapt-sampling,byrd}.
Alternative approaches implement Gaussian random walks \cite{gauss} and cyclical schedules \cite{cbs}, whilst others propose adaptations as a function of the local curvature in the loss landscape \cite{abs}.
A key distinction with Arbiter is that we train a neural network agent to \textit{learn} scheduling heuristics, based on gradients from a meta-objective function.
We nevertheless recognize the effectiveness of previous approaches, and foresee Arbiter as a source of complementary scheduling information.
We conduct a simple illustrative experiment in Section \textbf{\ref{sec:exp-bs}} to reinforce this prospect.
\\\\
\textbf{Hyperparameter optimization.}
Hyperparameters are typically optimized using grid- or random-search methods \cite{bergstra1} which perform well in many machine learning problems, but scale poorly 
to over-parameterized settings (i.e. DNNs). 
Contemporary work led to multi-armed bandit methods such as Hyperband \cite{hyperband} and successive-halving \cite{suc-halve}, in addition to model-based approaches such as Bayesian optimization \cite{snoek} and reinforcement learning \cite{zoph}, which search the hyperparameter space more efficiently, but inherit black-box properties that ignores structure in the objective.
Gradient-based methods were developed to access this structure, making them scalable to tune millions of hyperparameters within deep architectures \cite{maclaurin,pedrogosa}. Gradients have been implemented extensively within HO, including to learning rates \cite{hd,marthe,miccaelli}, regularization coefficients \cite{maclaurin,rtho,fran18}, and neural architecture search \cite{darts}. 
Another class of algorithms view the HO problem as a gray-box by considering the inner optimization structure, but relax the need for the meta-objective to be differentiable: population-based training \cite{pbt}, hypernetwork-based HO \cite{stn,d-stn}, and persistent evolution strategies (PES) \cite{pes} are examples of gray-box approaches.
Hyper-learning interpolates between gradient-based and black-box approaches, but presents different properties to gray-box methods: although hyper-learning requires the batch size to be differentiable to the objective, it reparameterizes it within the objective using differentiable samples from a neural network. Since HO occurs on the basis of information learned from meta-objective responses, hyper-learning avoids the need for unrolled optimization, an aspect inherent to gradient-based approaches.
\\\\
\textbf{Hypernetworks.}
The algorithmic structure for Arbiter closely resembles that of a hypernetwork \cite{schmid,ha}. Generally speaking, a hypernetwork is a function for predicting the weights of another neural network. Hypernetworks have been broadly applied, including to architecture search \cite{brock}, predicting CNN weights \cite{hyp-cnn}, meta-learning \cite{meta-hypernets}, and hyperparameter optimization \cite{d-stn,stn,pbm-ho}. 
We draw particular attention to \cite{lorr18}, who optimize hypernetworks using a novel \textit{hyper-training} procedure. Their goal was to predict optimal weights by conditioning the hypernetwork input on hyperparameters, creating an efficient process to jointly optimize the hyperparameters and hypernetwork weights.
In contrast to hypernetworks, the goal of hyper-learning is to predict optimal \textit{hyperparameter samples} for another network\footnote{This network may in fact be a hypernetwork, such that gradients from the objective target the learned hyperparameter distribution constructed by the agent.}. 
Since we maintain the inner network parameters, and instead learn a mapping to samples for the batch size using an outer objective, the neural network agent in Arbiter therefore does not represent a hypernetwork in the strict sense.
As a result, hyper-learning avoids the challenges of addressing the diversity of weight space in a hypernetwork, and is not restricted to optimizing outer variables that appear in the inner objective \cite{stn}, giving us access to tune the batch size and potentially other dynamical hyperparameters (e.g. learning rates).

\section{Arbiter}
In this section, we begin by motivating the batch size scheduling problem (Sect. \textbf{\ref{sec:mot}}). We then present Arbiter by describing its differentiable hyper-learning agent (Sect. \textbf{\ref{sec:nn}}), bilevel program to officiate response signals (Sect. \textbf{\ref{sec:bilevel}}), and simple parameterization scheme to sustain gradients (Sect. \textbf{\ref{sec:h}}).
To conclude, we summarize Arbiter's optimization process within a practical algorithm (Sect. \textbf{\ref{sec:alg}}).
\subsection{Motivation}
\label{sec:mot}
Consider a training dataset $\mathcal{D}:\lbrace (x_i, y_i)\rbrace$ of $M$ input-label pairs 
$\forall i \in \lbrace 1,...,M\rbrace$. 
In deep learning, we are interested in optimizing a set of neural network parameters $\mathbf{w}\in\mathbb{R}^d$, such that the objective function $L:\mathbb{R}^d\rightarrow\mathbb{R}$ is sufficiently minimized over all data-points. This is achieved by computing gradients $\mathbf{g}^{\mathcal{D}}\in\mathbb{R}^d$ to guide the parameters in the descent direction toward the minima.
Under mini-batch stochastic gradient descent (SGD), these gradients are averaged over mini-batches of data $\mathcal{B}\subset\mathcal{D}$ pulled randomly and without replacement from the dataset:
\begin{equation}
\mathbf{g}^{\mathcal{B}}(\mathbf{w})=\dfrac{1}{B}\sum_{j\in\mathcal{B}}\dfrac{\partial L(x_{j},y_{j})}{\partial\mathbf{w}}.
\label{eq:g}
\end{equation}
Crucially, the gradients in \eqref{eq:g} represent an approximation to the full-batch gradient (i.e. $\mathbf{g}^{\mathcal{B}}\approx\mathbf{g}^{\mathcal{D}}$), the quality of which dictates the amount of \textit{noise} in the gradients; poorer approximations equate to noisier gradients, and vice versa. 
Gradient noise sustains an important property when training deep networks: it facilitates an escape mechanism, where parameters are given energy to move from sharp regions in the objective landscape and into flatter minima \cite{onrelation}. 
Strong theoretical and empirical evidence has shown this mechanism to be intrinsically linked to improved generalization \cite{break-even,smith-sgd,nips-sgd,three-factors}. 
Most importantly for our discussion, gradient noise is directly influenced by the \textit{batch size} ($B$) due to its control over the mini-batch statistics and parameterization within the objective (Eq. \ref{eq:g}), where smaller batch sizes induce greater noise \cite{bayes-sgd,smith-sgd}.
For this reason, it is common practice to initialize training with a large learning rate to batch size ratio \cite{three-factors,nips-sgd}, followed by a subsequent tempering of the noise to allow the parameters to converge. 
\\\\
Within the context of batch size scheduling, we should therefore select a \textquoteleft sufficiently' small initial batch size, and choose a scheduling heuristic for increasing the batch size as training progresses. However, these are difficult choices to make \textit{a priori} due to the non-convex optimization and vast dimensionality of the objective landscape sustained by neural networks \cite{goldilocks}, which makes them highly sensitive to batch size choices \cite{bayes-sgd,how-sgd}.
At the same time, if the batch size is improperly set, we risk imposing suboptimal gradient noise to the parameters; 
this may have particularly deleterious consequences during early-stage iterations, which are crucial for determining the remaining optimization trajectory \cite{break-even}.
To address this problem, we therefore require a principled approach to interact with the dynamics of our parameters for receiving guidance to select optimal batch sizes throughout training.

\subsection{Hyper-learning agent} 
\label{sec:nn}
%
%
Let $
Z=\mathbb{N}
$ 
represent the set of all possible batch sizes\footnote{The maximum batch size in this set is determined by the user's GPU capacity. In our work, we were limited to batch size of 600 for the experiments.}.
We are proposing to search for optimal values in this set as a function of an evolving inner dynamical system ($\mathbf{w}$) using a differentiable \textit{hyper-learning} agent. We formulate the agent as a neural network $\boldsymbol{\phi}\in\mathbb{R}^p$ to perform a mapping $\boldsymbol{\phi}:\mathcal{X}\rightarrow \mathcal{S}$ from a given mini-batch of data 
$\mathcal{X}\in\mathbb{R}^{B\times D}$ to samples $\mathcal{S}\in\mathbb{R}^N$, representing $N$ batch size predictions. 
In this respect, the agent assumes the role of a \textit{hyperparameter generator}, tasked with generating a
set of new batch sizes around the current batch size $B\in Z$. 
\begin{figure*}[t!]
\begin{center}
\includegraphics[width=\textwidth]{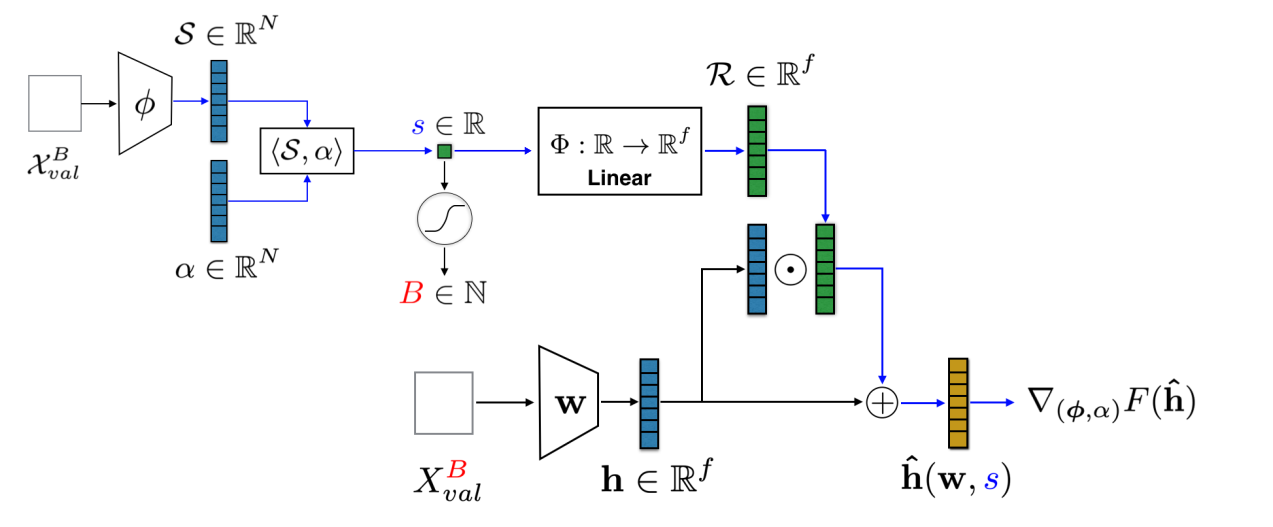} 
\end{center}
\caption{\textbf{Architecture for Arbiter.} \textit{Generating batch size samples.} Given a mini-batch of validation data $\mathcal{X}^{B}_{val}$, the agent $\boldsymbol{\phi}:\mathcal{X}\rightarrow\mathcal{S}$ learns to map the data to batch size samples ($\mathcal{S}$) in a neighborhood around the current batch size ($B$). Samples are assigned probabilities by $\alpha$, and mixed to yield a representative sample ($\textcolor{blue}{s}$), which is mapped through a sigmoid function ($\sigma$) to obtain the batch size in hyperparameter form ($\textcolor{red}{B}$). 
\textit{Parameterizing the high-level features.} We perform a linear mapping $\Phi:\mathbb{R}\rightarrow\mathbb{R}^f$ to project the mixed sample ($\textcolor{blue}{s}$) to the same $f-$dimensional subspace as the high-level features ($\mathbf{h}$) for parameterization. Intuitively, this represents perturbations to the input data statistics caused by the batch size. This creates a response function that we can differentiate over, without needing to backpropagate through the inner system ($\mathbf{w}$) to the hyperparameter itself ($\textcolor{red}{B}$), but rather to its differentiable proxy ($\textcolor{blue}{s}$). The \textcolor{blue}{blue arrows} indicate the forward path, which we subsequently backpropagate along to update Arbiter ($\boldsymbol{\phi}, \alpha)$. 
}
\label{fig:phi}
\end{figure*}
\\\\
The idea to explore the hyperparameter space using a neural network is not new, and in fact forms the basis of population-based HO \cite{pbm-ho}. The difference with our approach is that we do not require a group of clones for $\mathbf{w}$
to inherit separate hyperparameter values, and instead elect a cheaper neural network to \textit{predict optimal samples} from the space. This formulation avoids the need to train and subsequently cull multiple redundant networks using a mutation policy.
\\\\
\textbf{Continuous relaxation.} 
However, evaluating $N$ samples at each optimization step would re-create an inefficient \textquoteleft black-box'-styled search. 
We instead draw inspiration from \cite{darts} and relax the search by introducing parameters $\alpha\in\mathbb{R}^N$ into the agent's architecture
Letting $S_{i}\in\mathcal{S}$ and $\alpha_{i}\in\alpha$, this enables us to assign separate softmax weights to the samples and subsequently mixing as follows:
\begin{equation}
s = \sum_{i=1}^{N}\dfrac{\exp(\alpha_i)}{\sum_{j=1}^N \exp(\alpha_j)} S_i
\label{eq:s}
\end{equation}
to yield a single representative sample for the batch size $s\in\mathbb{R}$, as illustrated in Figure \textbf{\ref{fig:phi}}. 
With this setup, the agent is free to construct an implicit probability distribution 
over the batch size $p_{\boldsymbol{\phi}}(B|\mathcal{X}$), from which we can retrieve samples $\mathcal{S}\thicksim p_{\boldsymbol{\phi}}(B|\mathcal{X}$), leaving $\alpha$ to identify those   most likely to improve generalization.
\subsection{Optimization}
\label{sec:bilevel}
To drive the searching process,
we require a mechanism for interacting with the inner dynamics through our predicted batch size samples.
Our goal in hyper-learning is to predict hyperparameter samples to optimize the dynamics of another neural network.
In other words, we seek to optimize the batch size predicted by our agent, subject to the optimality of the inner parameters.
This describes a bilevel optimization problem \cite{colson}:
\begin{align}
s^* = 
\textup{min}_{(\boldsymbol{\phi},\alpha)}\ F(\mathbf{w^*},s) \nonumber \\
\textup{s.t.}\ \ \mathbf{w^*} = \textup{argmin}_{\mathbf{w}}\ L(\mathbf{w}, B)
\label{eq:bilevel}
\end{align}
where $F:\mathbb{R}^d\rightarrow\mathbb{R}$ denotes a \textit{meta-} (or outer) objective over validation data, 
$L:\mathbb{R}^d\rightarrow\mathbb{R}$ an \textit{inner} objective over training data, and $s^*\in\mathbb{R}$ the optimal batch size.
\\\\
In the context of HO, $\mathbf{w^*}$ represents a unique solution for optimal weights that is difficult to obtain in closed-form \cite{stn}. 
We follow previous work by approximating $\mathbf{w^*}$ with $\mathbf{w}_T$, the weights after $[T]=\lbrace1,...,T \rbrace$ gradient descent steps \cite{maclaurin,rtho,fran18}. 
In practice, this approach is challenged by the need to unroll all optimization states for computing $\partial\mathbf{w}_t/\partial\boldsymbol{\lambda}, \forall t\in[T-1]\rightarrow0$ for a vector of hyperparameters $\boldsymbol{\lambda}\in\mathbb{R}^l$, which often creates a non-smooth meta-loss landscape ubiquitous with severe gradient variance, leading to known pathology \cite{pes}.
\\\\
Based on Eq.\eqref{eq:bilevel}, we can see that we avoid this problem by using the continuous batch size ($s$) as a proxy for the discrete batch size ($B$) in the meta-objective. 
The solution to the above program is found by minimizing Arbiter's parameters $\lbrace\boldsymbol{\phi},\alpha\rbrace_{t}$ in parallel to the inner optimization procedure $\forall t\in[T]$, and using the inner weights $\mathbf{w}_{t+1}$ 
to compute responses to the predicted batch size ($s_t$).
Although this incurs additional compute cost over normal DNN training, we do not require unrolling the computational graph to obtain gradients for the batch size, because we instead leverage the response information \textit{learned} by Arbiter while tuning its proxy.
Recalling the relaxation scheme in Eq.\eqref{eq:s}, we can therefore approximate the solution $s^*$ by computing $b_{T} = \textup{argmax}_{\alpha}(\mathcal{S}_{T})$ to select the \textquoteleft best' sample (i.e. highest probability) at $t=T$, and subsequently replace the current batch size with $B=\sigma(b_{T})$ by mapping the best sample to a natural number through a sigmoid function (Fig. \textbf{\ref{fig:phi}}).

\subsection{Parameterizing the high-level features}
\label{sec:h}
In Section \textbf{\ref{sec:nn}}, we presented a strategy to bypass the need for Monte Carlo gradient estimation \cite{mcgrads} for the batch size by using an implicit probability distribution $p_{\boldsymbol{\phi}}(B|\mathcal{X}$). We then developed a bilevel optimization procedure in the previous section for moving samples from this distribution toward optimality. However, the batch size possesses no differentiable structure within the inner network, meaning we do not have a function for reaching the samples with gradients from the meta-objective in Eq.\eqref{eq:bilevel}.
\\\\
\textbf{Linear mapping to feature space.} 
Inspired by previous works \cite{stn}, we propose 
the use of a linear gating mechanism to approximate the \textquoteleft best-response' of the learnable features to the batch size. 
More specifically, we use a linear mapping $\Phi:s\rightarrow\mathcal{R}$ to project the reparameterized batch size lying in $\mathbb{R}$ to the subspace occupied by the high-level features $\mathbf{h}\in\mathbb{R}^f$ of the inner network, and use its representation $\mathcal{R}\in\mathbb{R}^f$ to gate the learned high-level features from the inner network. This generates an additional feature term, as illustrated in Figure \textbf{\ref{fig:phi}}. For a given predicted batch size $s$, and linear weights $\mathbf{A}_{\Phi}\in\mathbb{R}^{f}$, we can describe this process by the following affine transformation:
\begin{equation}
\mathbf{\hat{h}}(\mathbf{w},s) := \mathbf{A}_{\Phi}(s)\odot \mathbf{h} + \mathbf{h},
\label{eq:h}
\end{equation}
which corresponds to the \textit{response} of the features to the predicted batch size.
\\\\
This is a reasonable approximation to make for three reasons. First, we know that the batch size controls the distribution of our mini-batch data statistics, which can be represented by the high-level features extracted by $\mathbf{w}$ \cite{improved-gans}; we therefore expect a commensurate response from the features to the predicted batch size. Second, we know that the non-linearities in the data have been captured by the parameters in the preceding layers of $\mathbf{w}$, enabling us to fit linear functions within this lower-dimensional manifold. Third, according to \cite{stn}, the best-response function for a linear network can be represented by such a gating architecture.
Notably, in hyper-learning we do not use the best-response to learn network weights directly as with hypernetworks \cite{lorr18}. Instead, this is applied to approximate the response of the hidden units (i.e. features) to the batch size, making it simpler in execution. 
\\\\
The introduction of this mechanism allows us to parameterize the features, and update Arbiter's weights through back-propagation from the meta-objective:
%
\begin{equation}
\nabla_{(\boldsymbol{\phi},\alpha)}F(\mathbf{\hat{h}})
:=
\dfrac{\partial F}{\partial \mathbf{\hat{h}}}\dfrac{\partial \mathbf{\hat{h}}}{\partial \boldsymbol{\phi}} +
\dfrac{\partial F}{\partial \mathbf{\hat{h}}}\dfrac{\partial \mathbf{\hat{h}}}{\partial \alpha}.
\label{eq:hgrad}
\end{equation}
Naturally, one may question the sufficiency of the linear mapping to preserve the structure of the batch size within the higher-dimensional subspace. We argue that the exact representation of the batch size is irrelevant,
as long as the agent learns to distinguish the effects imposed by small and large batch sizes on the high-level features and meta-objective; this argument has been made for similar mapping operations \cite{sci-net}.
We appeal to the experimental results as an initial justification for this assumption, and leave a deeper theoretical analysis to future work. 

\subsection{Training algorithm}
\label{sec:alg}
We summarize our developments within a practical training algorithm (Alg. \textbf{\ref{alg}}). In short, Arbiter optimizes its predicted samples relative to meta-objective responses from one-step inner weights ($\mathbf{w}_{t+1}$), and searches for an optimal batch size over $T$ iterations. 
We are currently unaware of the convergence guarantees for our algorithm, but experimentally Arbiter is capable of reaching stable solutions across various tasks. 
\begin{wrapfigure}{R}{0.5\textwidth}
\begin{minipage}{0.5\textwidth}
\vspace{-15mm}
    \begin{algorithm}[H]
    \caption{Batch size adaptation with Arbiter. The symbols ($\sigma,\psi$) represent the sigmoid and softmax functions, respectively.}
    \label{alg}
        \begin{algorithmic}
          \STATE {\bfseries Initialize:} Meta-parameters: $N, N_{learn}, \zeta_{\phi}, \zeta_{\alpha}$
          \STATE {\bfseries Initialize:} Arbiter: $\lbrace\boldsymbol{\phi}\in\mathbb{R}^p,\ \alpha\in\mathbb{R}^N\rbrace$
          \FOR{$n=1$ {\bfseries to} $N_{epochs}$}
          \FOR{$t=0$ {\bfseries to} $T_{train}$}
          \STATE 
        $\mathbf{w}_{t+1}\leftarrow\mathbf{w}_t -\eta\nabla_{\mathbf{w}}L(\mathbf{w}_t , B_{n})$ 
          \STATE 
          $\mathcal{S}\thicksim p_{\phi}(B_{n}|\mathcal{X})$ , $s_{t}=\textup{dot}\big(\mathcal{S},\psi(\alpha)\big)$ 
          \STATE 
          $\boldsymbol{\phi}_{t+1}\leftarrow\boldsymbol{\phi}_t - \zeta_{\phi}\nabla_{\phi}F\big(\mathbf{\hat{h}}(\mathbf{w}_{t+1}, s_{t})\big)$
          \STATE 
        $\alpha_{t+1}\leftarrow\alpha_t - \zeta_{\alpha}\nabla_{\alpha}F\big(\mathbf{\hat{h}}(\mathbf{w}_{t+1}, s_{t})\big)$
          \ENDFOR
          \IF{$n=N_{learn}$}
          \STATE Update batch size:\\ \hspace{5mm} $B_{n+1} = \sigma\big(\textup{argmax}_{\alpha}(\mathcal{S}_{T})\big)$
          \STATE Reset weights:\\ \hspace{5mm} $\alpha\thicksim\mathcal{N}(0,1)\in\mathbb{R}^N$
          \ENDIF
          \ENDFOR
        \end{algorithmic}
    \end{algorithm}
\end{minipage}
\vspace{-10mm}
\end{wrapfigure}

\newpage
\section{Experiments}
\label{sec:exp}
We conduct several representative experiments. First, we examine 
Arbiter's performance as a stand-alone batch size scheduler under various dynamical conditions (Sect. \textbf{\ref{sec:exp-optim}}). We then illustrate the benefits of batch size exploration with Arbiter in combination with fixed batch size schedules (Sect. \textbf{\ref{sec:exp-bs}}). Lastly, we study the impact of Arbiter's control over gradient variance on hypergradient descent of the learning rate \cite{hd} in Section \textbf{\ref{sec:exp-lr}}.
 

\subsection{Batch size scheduling}
\label{sec:exp-optim}
\begin{figure*}[h!]
\begin{center}
    \begin{subfigure}[t]{0.23\textwidth}
        \includegraphics[width=\textwidth]{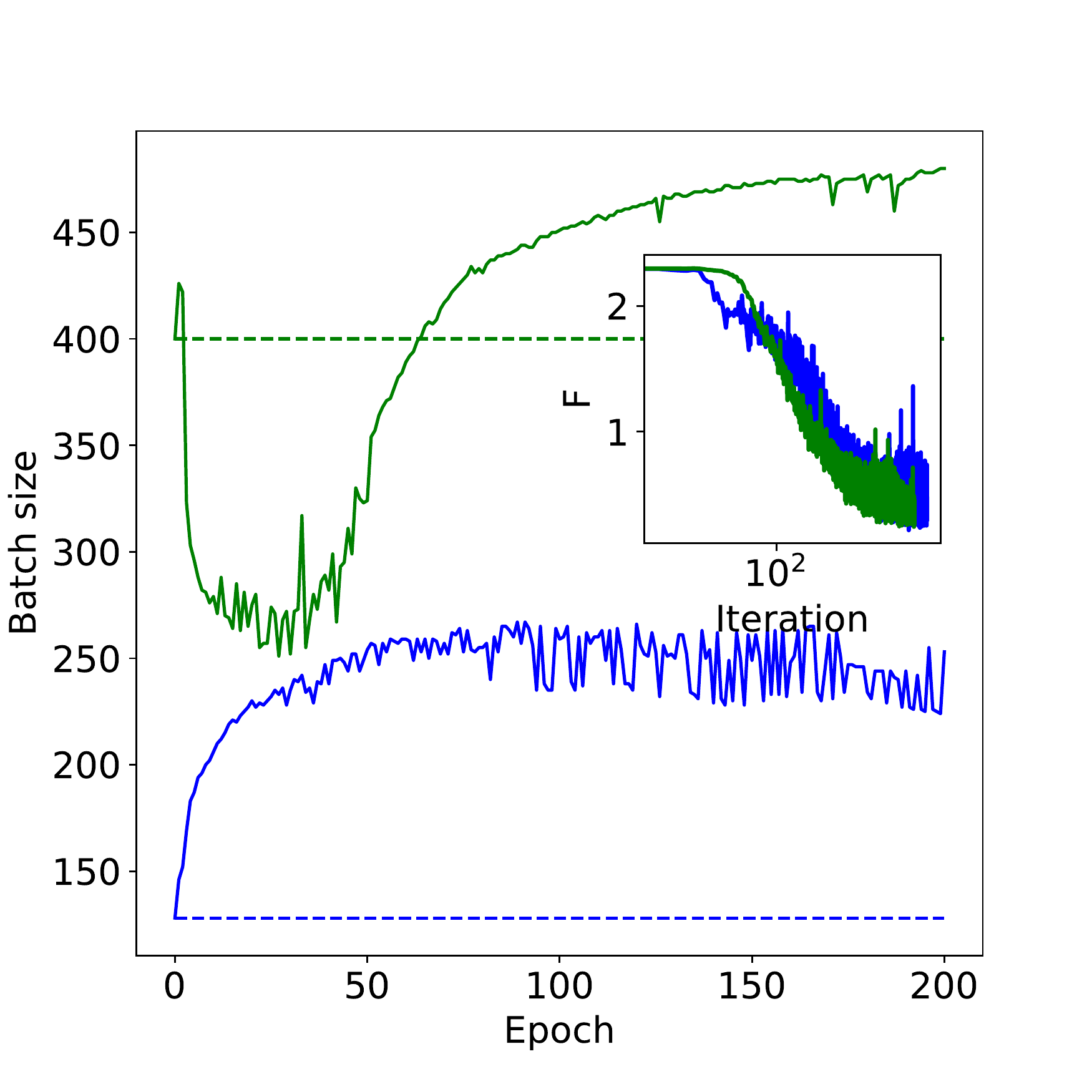} 
    \end{subfigure}
    \begin{subfigure}[t]{0.23\textwidth}
        \includegraphics[width=\textwidth]{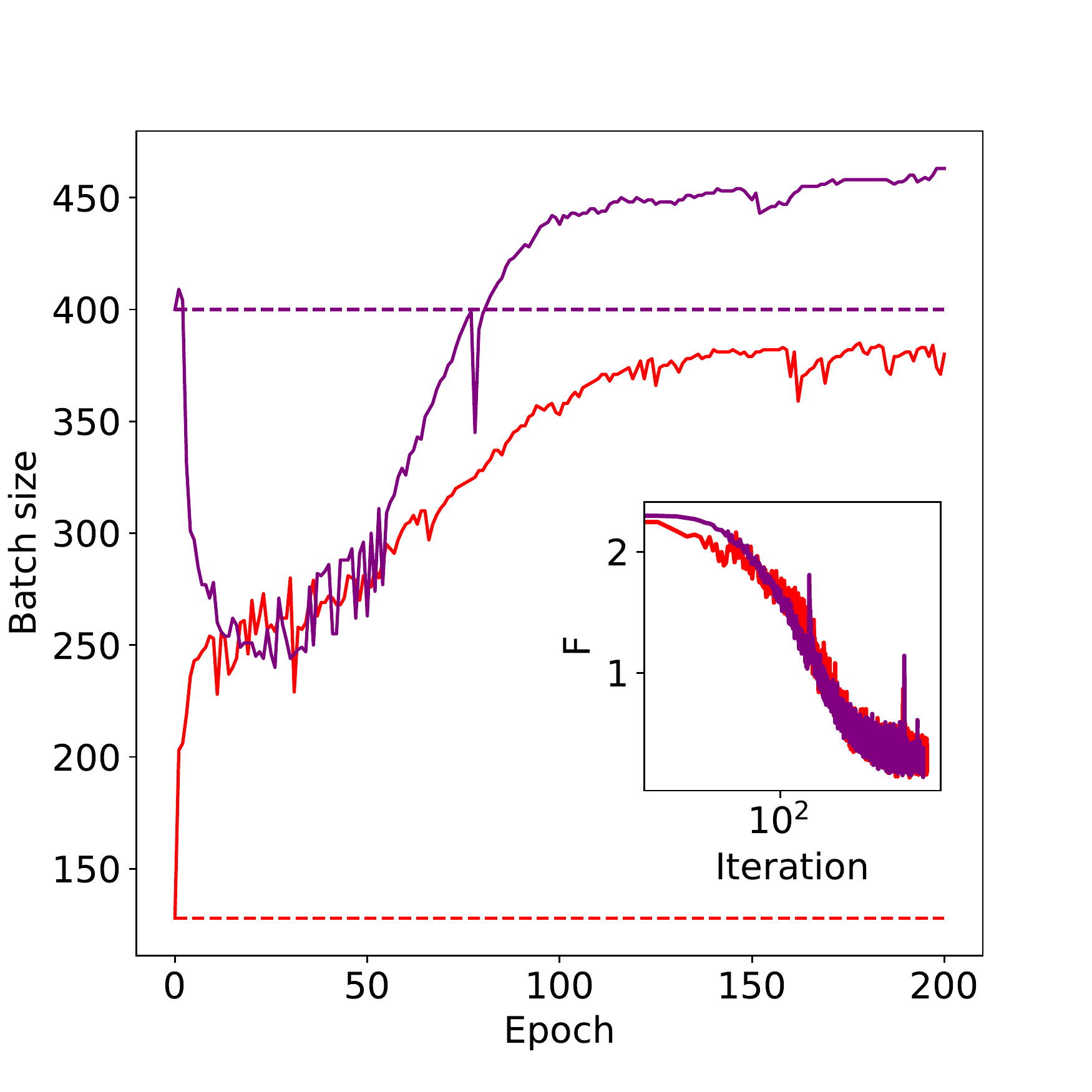} 
    \end{subfigure}
    \begin{subfigure}[t]{0.23\textwidth}
        \includegraphics[width=\textwidth]{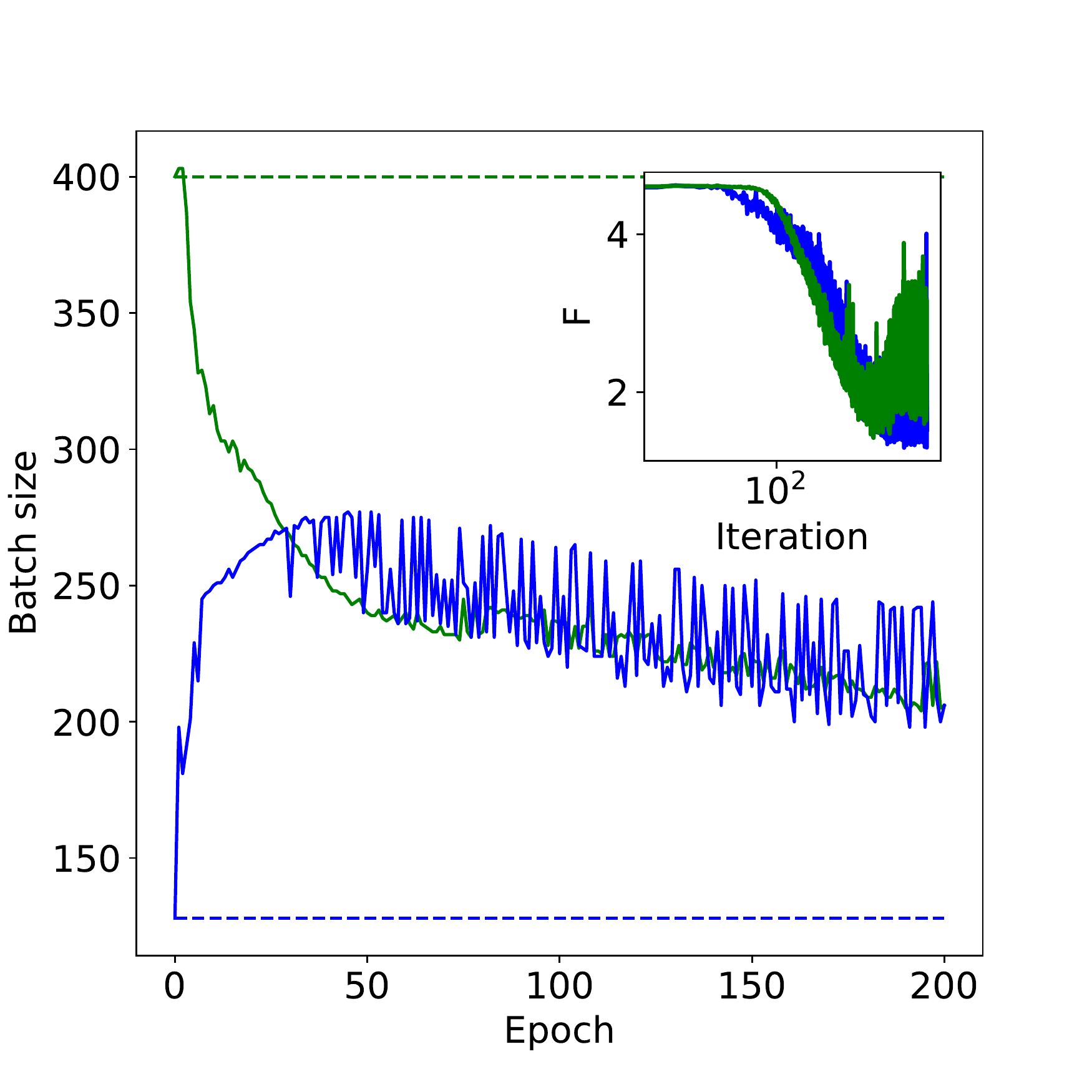} 
    \end{subfigure}
    \begin{subfigure}[t]{0.23\textwidth}
        \includegraphics[width=\textwidth]{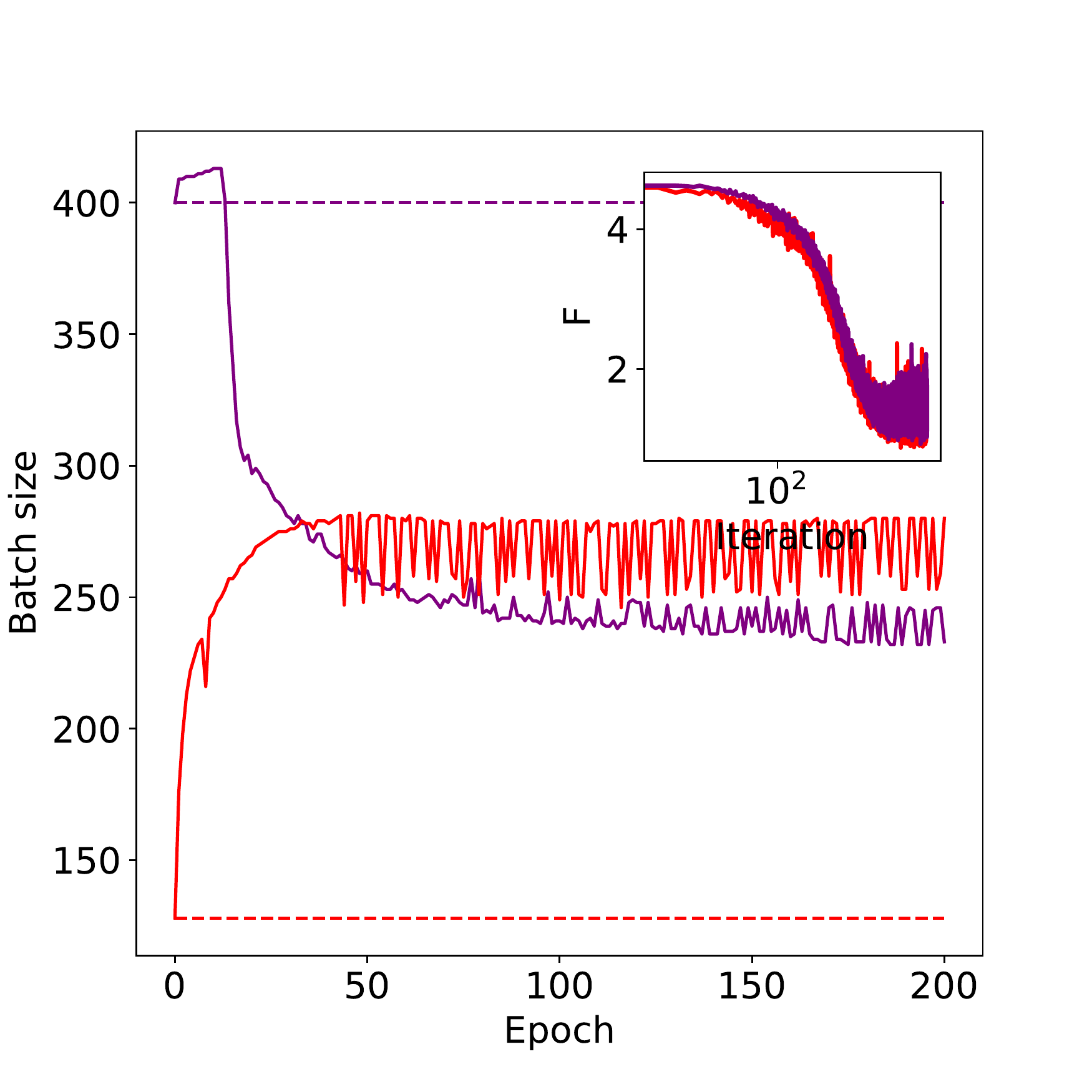} \\ 
        \vspace{-15mm}
    \end{subfigure} 
    %
    %
    \begin{subfigure}[t]{0.23\textwidth}
        \includegraphics[width=\textwidth]{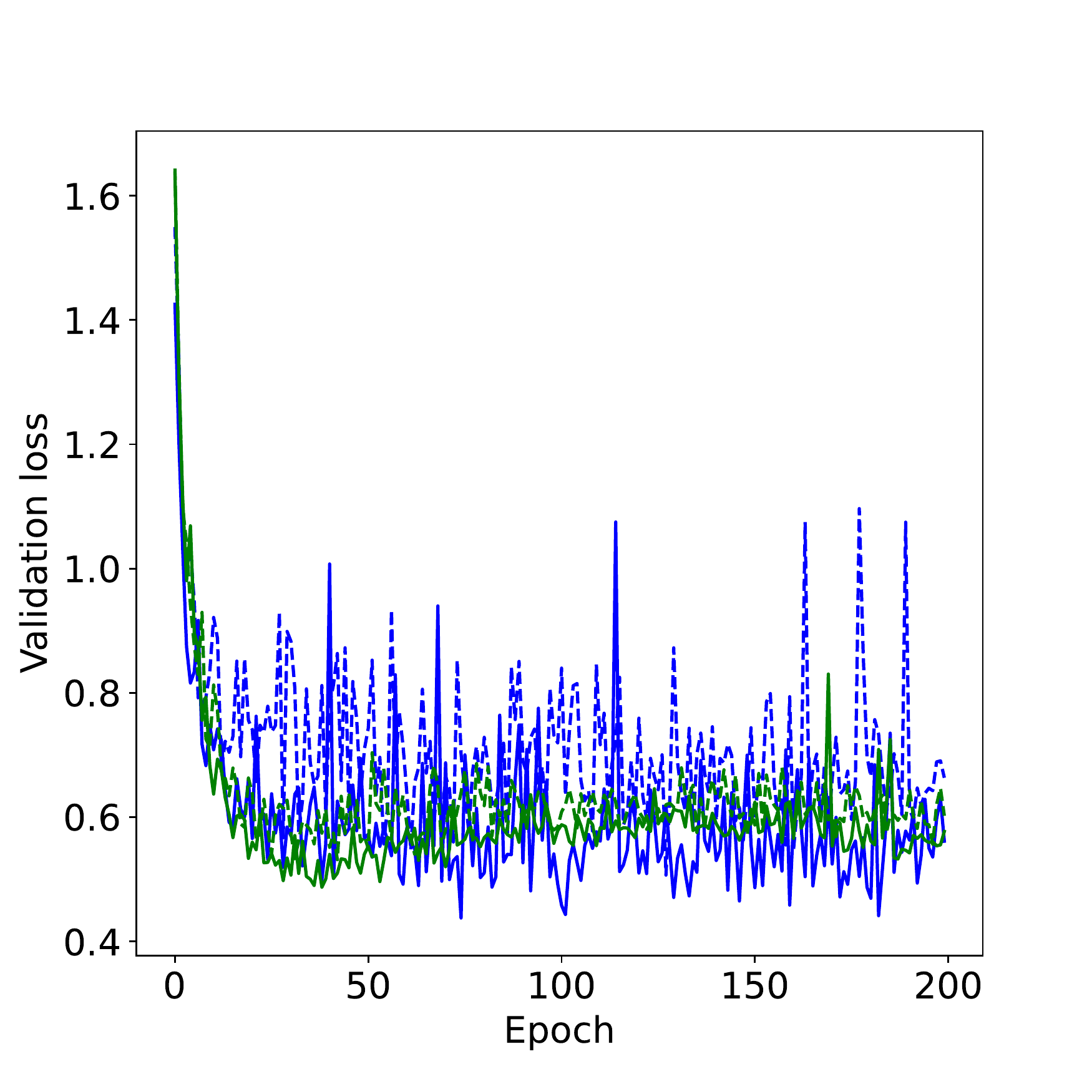} 
        \vspace{-6.5mm}
        \caption*{\hspace{30mm}(a)}
    \end{subfigure}
    \begin{subfigure}[t]{0.23\textwidth}
        \includegraphics[width=\textwidth]{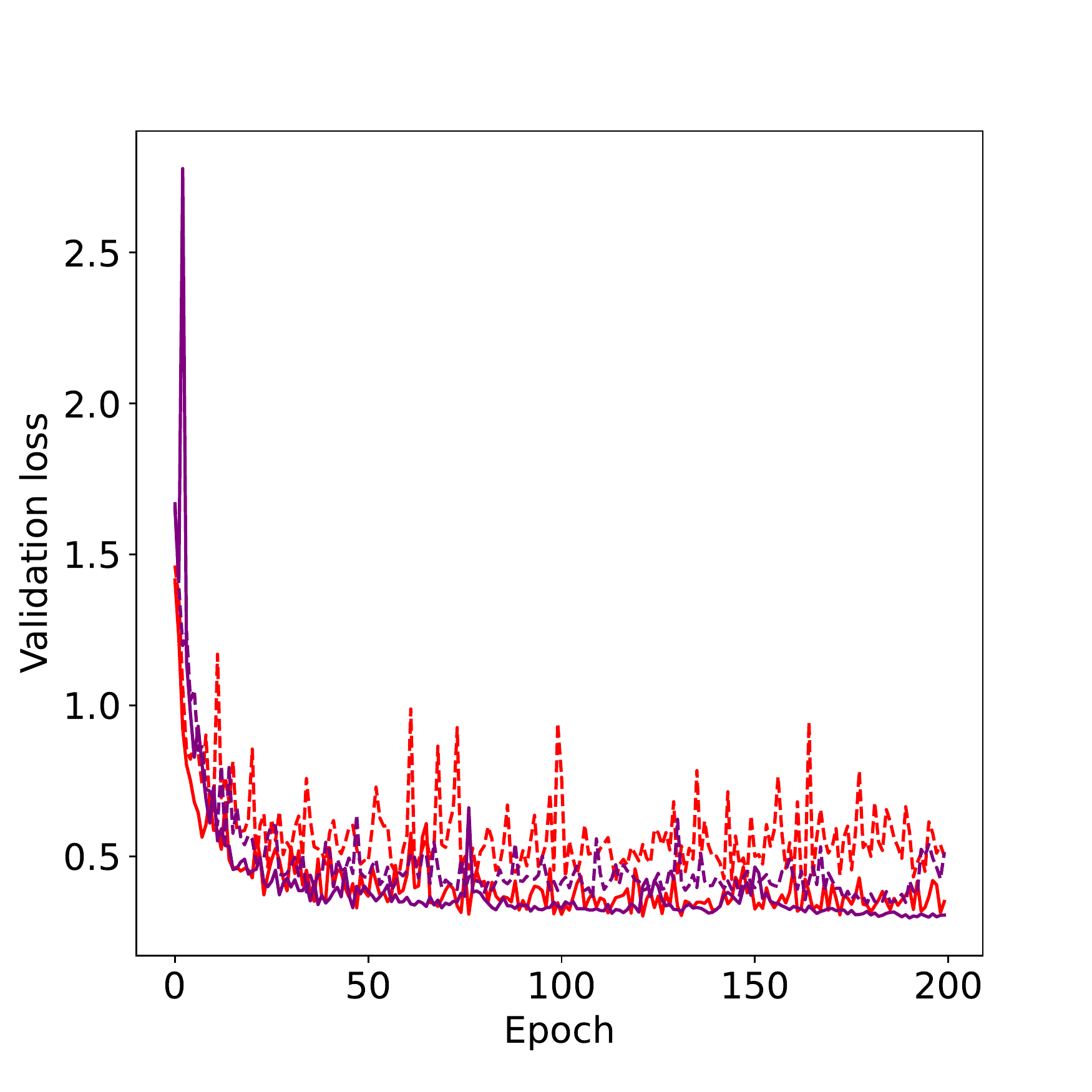} 
    \end{subfigure}
    \begin{subfigure}[t]{0.23\textwidth}
        \includegraphics[width=\textwidth]{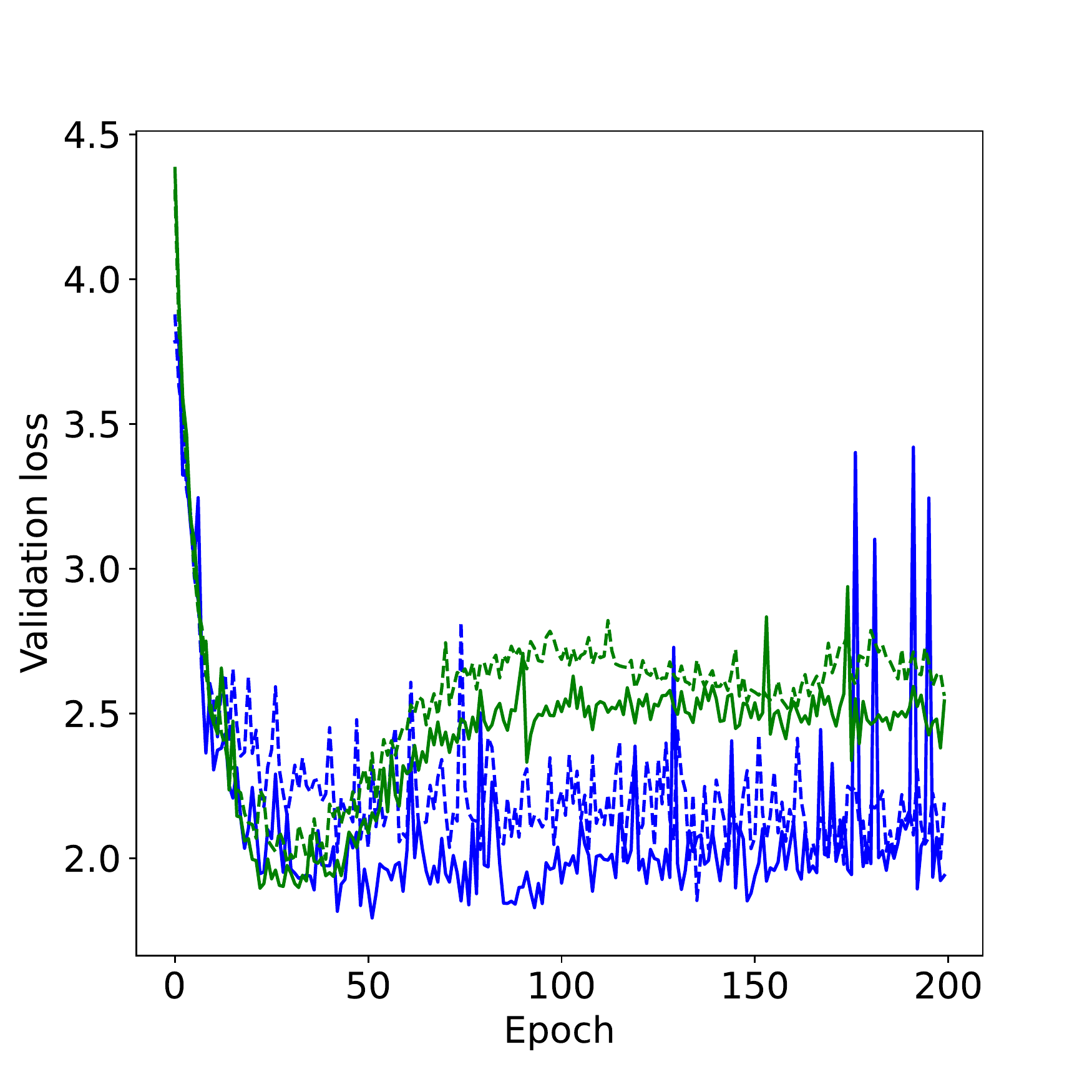}
        \vspace{-6.5mm}
        \caption*{\hspace{30mm}(b)}
    \end{subfigure}
    \begin{subfigure}[t]{0.23\textwidth}
        \includegraphics[width=\textwidth]{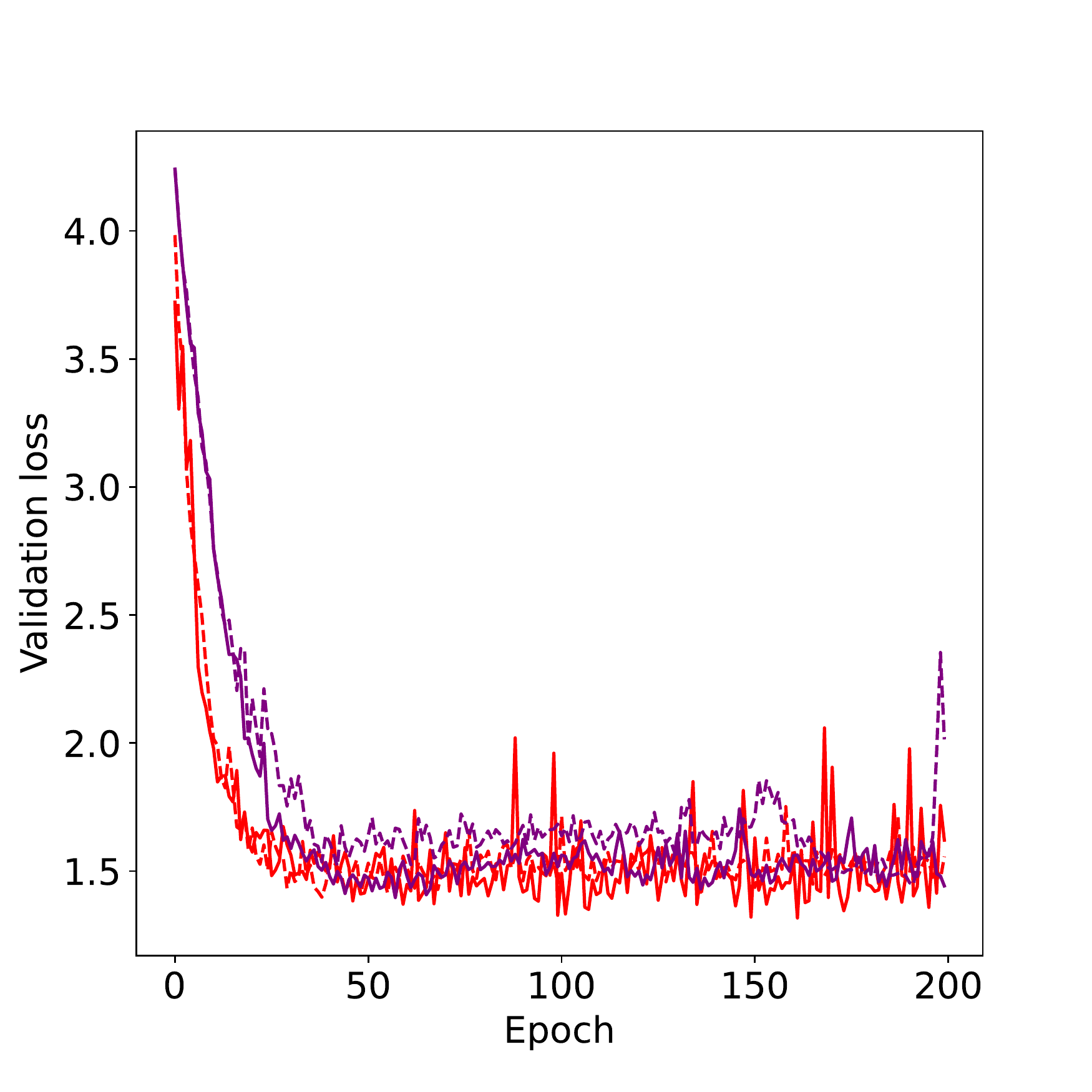} \\ 
    \end{subfigure} 
\end{center}
\vspace{-2.5mm}
\caption{
\textbf{Loss curves and batch size schedules determined by Arbiter}, for the experiments described in Sect. \textbf{\ref{sec:exp-optim}}.
(a) Results on CIFAR-10, using the VGG-11 (\textbf{left}) and Wide ResNet-16-4 (\textbf{right}) architectures. (b) Results on CIFAR-100 for VGG-11 (\textbf{left}) and Wide ResNet-16-4 (\textbf{right}). 
In this experiment, Arbiter was tasked with adapting the batch size to accommodate for initially stochastic (\textcolor{blue}{blue} and \textcolor{red}{red}) and non-stochastic dynamics (\textcolor{green}{green} and \textcolor{purple}{purple}). We observe that Arbiter's batch size adaptations (solid lines) reflect a decreasing meta-objective ($F$, inset), leading to schedules that yield lower validation error in all experiments compared to constant batch sizes (dashed lines).
}
\label{fig:exp-optim}
\end{figure*}

In this section, we show that Arbiter can compensate for initially deleterious dynamics, and perform effective subsequent batch size scheduling, purely from gradients.
We experiment on both the CIFAR-10 and CIFAR-100 datasets using two modern architectures (VGG-11 and Wide ResNet-16-4) under stochastic and non-stochastic dynamics. For stochastic dynamics, we initialized training with a learning rate of $\eta=0.1$ and a batch size of $B=128$. For non-stochastic dynamics, we used $\eta=0.01$ and $B=400$. In both scenarios, Arbiter was implemented to correct the dynamics by adapting the batch size every epoch (i.e. $N_{learn}=1$).
\\\\
From our knowledge of the learning dynamics, we expect the gradients to inform Arbiter to increase the batch size amidst noisy dynamics, and decrease the batch size to introduce greater noise into non-stochastic dynamics. In Figure \textbf{\ref{fig:exp-optim}}, we observe that Arbiter responds correctly in both cases. By continuing to follow the meta-objective gradients, Arbiter learns scheduling heuristics that lower the validation loss in all experiments compared to training with a constant batch size. 
%

\subsection{Flexibility within fixed heuristics}
\label{sec:exp-bs}
\begin{figure}[h!]
\begin{center}
    \begin{subfigure}[t]{0.3\textwidth}
        \includegraphics[width=\textwidth,height=0.78\textwidth]{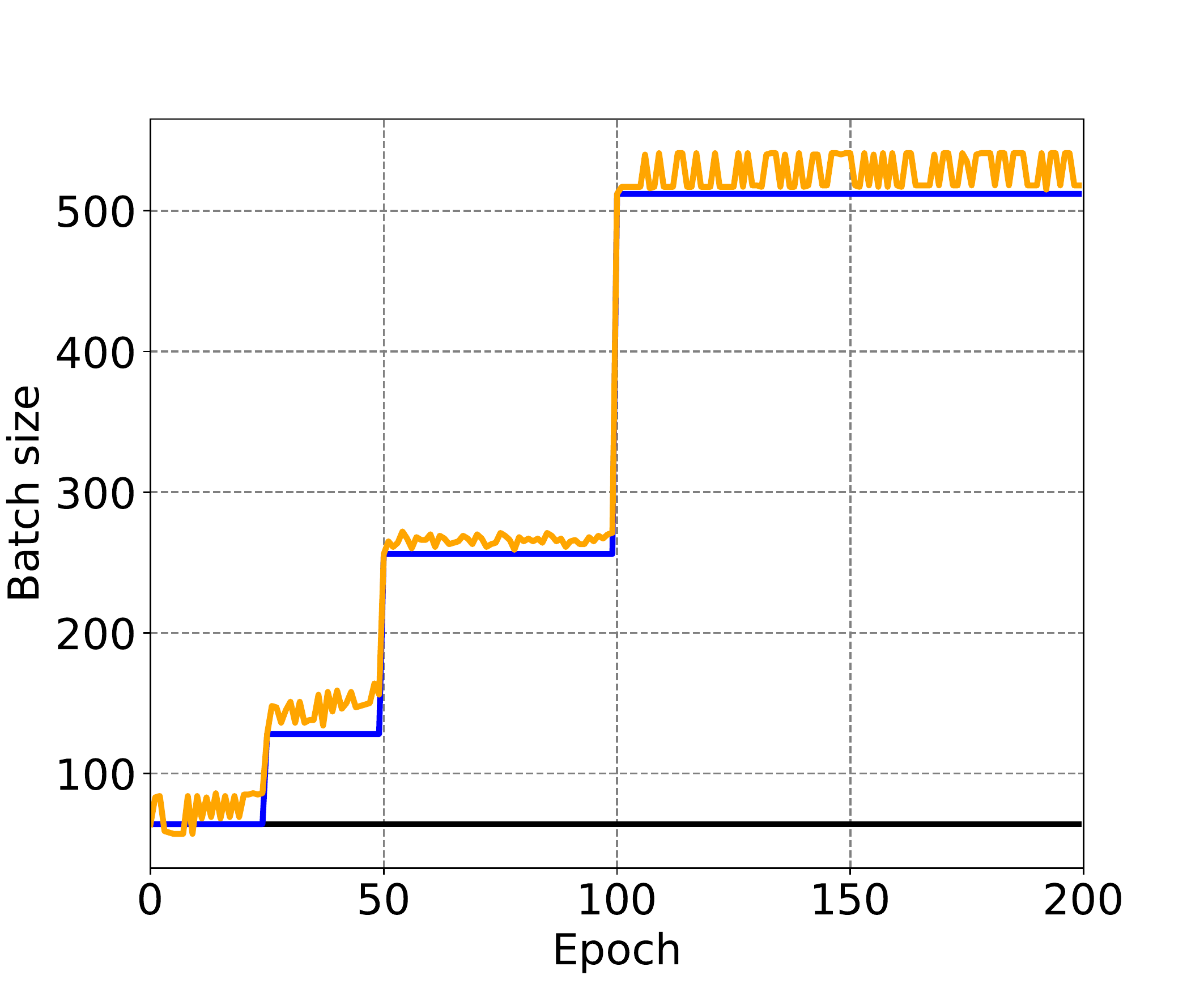} 
    \end{subfigure}
    \begin{subfigure}[t]{0.3\textwidth}
        \includegraphics[width=\textwidth,height=0.78\textwidth]{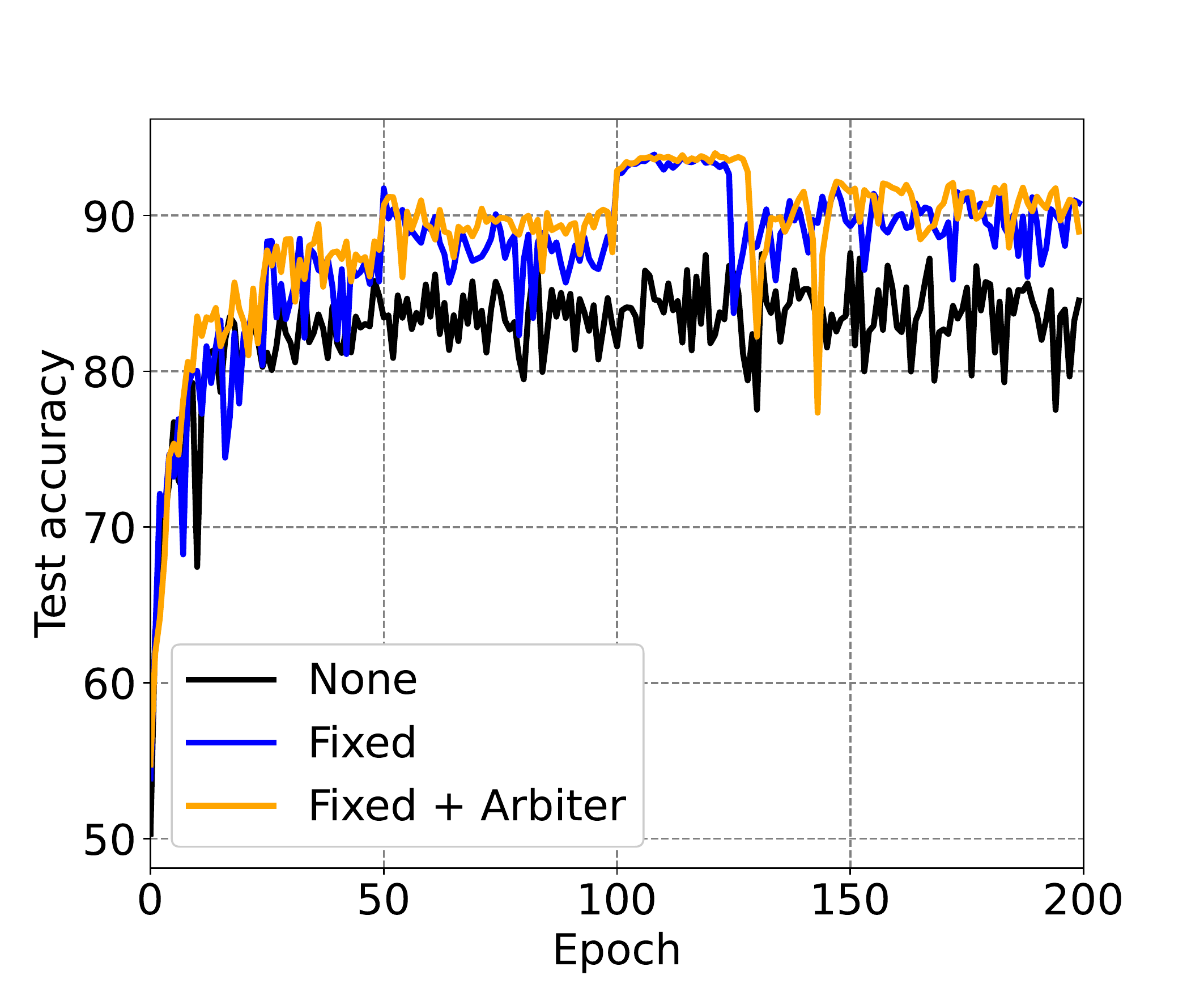}
    \end{subfigure}
\end{center}
\vspace{-2.5mm}
\caption{\textbf{Searching locally for more optimal batch sizes}, for the experiments in Sect. \textbf{\ref{sec:exp-bs}}. We observe that Arbiter selects different batch sizes from the pre-defined heuristic, yielding a schedule with higher test accuracy on CIFAR-10. 
}
\label{fig:exp-fixed}
\end{figure}

In this experiment, we demonstrate that Arbiter can be used to complement fixed batch size schedulers with greater flexibility.
By allowing a fixed heuristic to dictate scheduling, we give Arbiter freedom to search for locally \textquoteleft better' batch sizes and examine the resulting effects on generalization. We trained for 200 epochs on CIFAR-10 with the Wide ResNet-16-4 network, using a constant learning rate of $\eta=0.05$. The results are presented in Figure \textbf{\ref{fig:exp-fixed}}.
\\\\
We designed the following fixed batch size schedule: beginning with a batch size of $B_{0}=64$, we propose to increase the batch size to 128, 256, and 512 at epoch milestones of 25, 50, and 100, respectively. We then implemented Arbiter to conduct a local search between these milestones. From the results in Figure \textbf{\ref{fig:exp-fixed}}, we observe that Arbiter selects initially noisier batch sizes, before determining larger values for the remainder of training. As a result, we achieve higher test performance. 
%

%
%

\subsection{Variance reduction in stochastic meta-optimization}
\label{sec:exp-lr}
Recently it was shown that stochastic meta-optimization of the learning rate incurs short-horizon (or truncation) bias \cite{short-horizon,pes}, 
alluding to the issue with \textquoteleft greedily' 
lowering the learning rate in response to a rapidly decreasing meta-objective. 
As a result, the learning rate converges to near negligible values early in training, eliminating noise in the dynamics needed to escape poor local minima.  
The intrinsic relationship between the learning rate and batch size \cite{three-factors} motivated us to investigate whether Arbiter could help alleviate this issue by inducing  variance reduction via batch size adaptations \cite{short-horizon}.
\begin{figure}[h!]
\begin{center}
    \begin{subfigure}[t]{0.234\textwidth}
        \includegraphics[width=\textwidth]{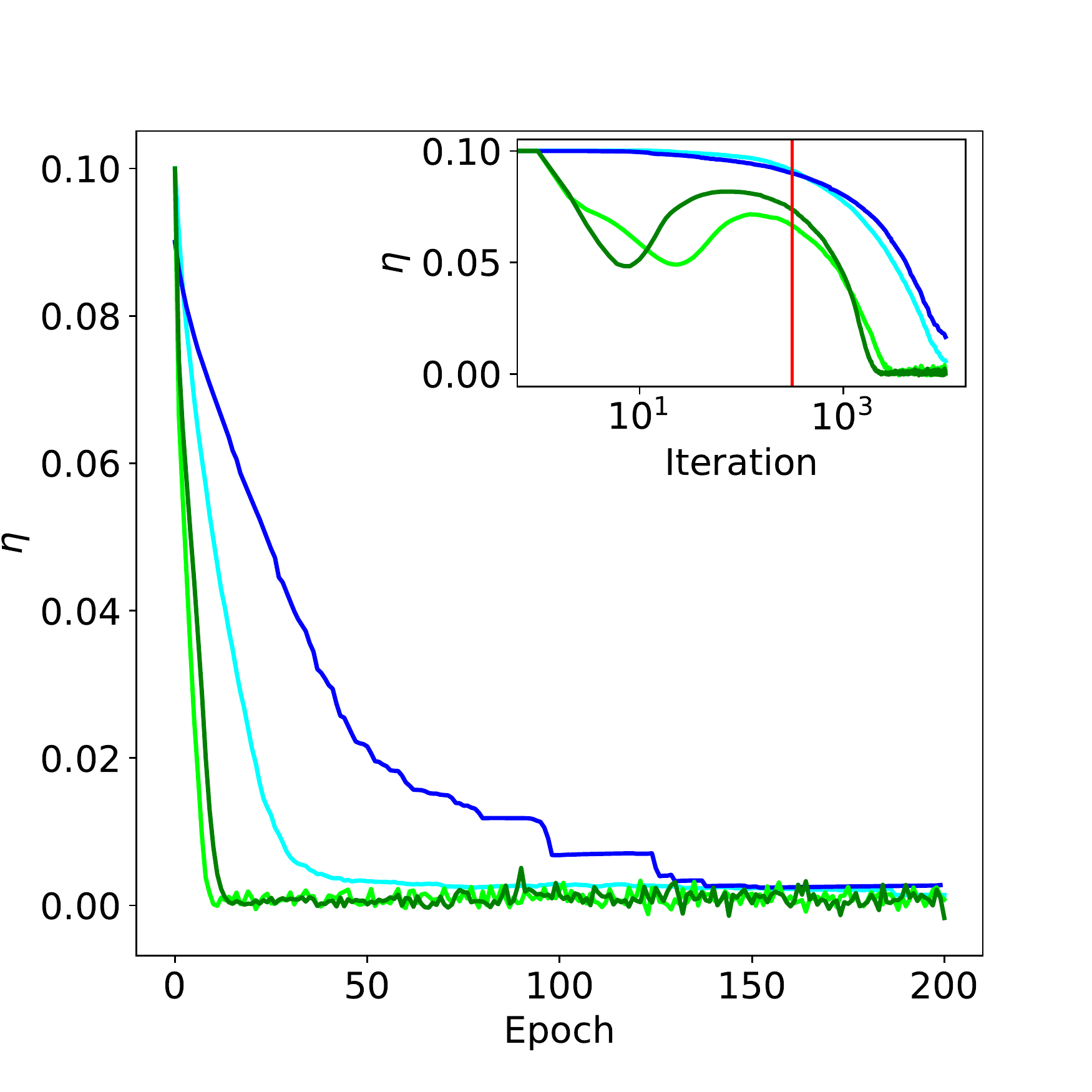} 
    \end{subfigure}
    \begin{subfigure}[t]{0.234\textwidth}
        \includegraphics[width=\textwidth]{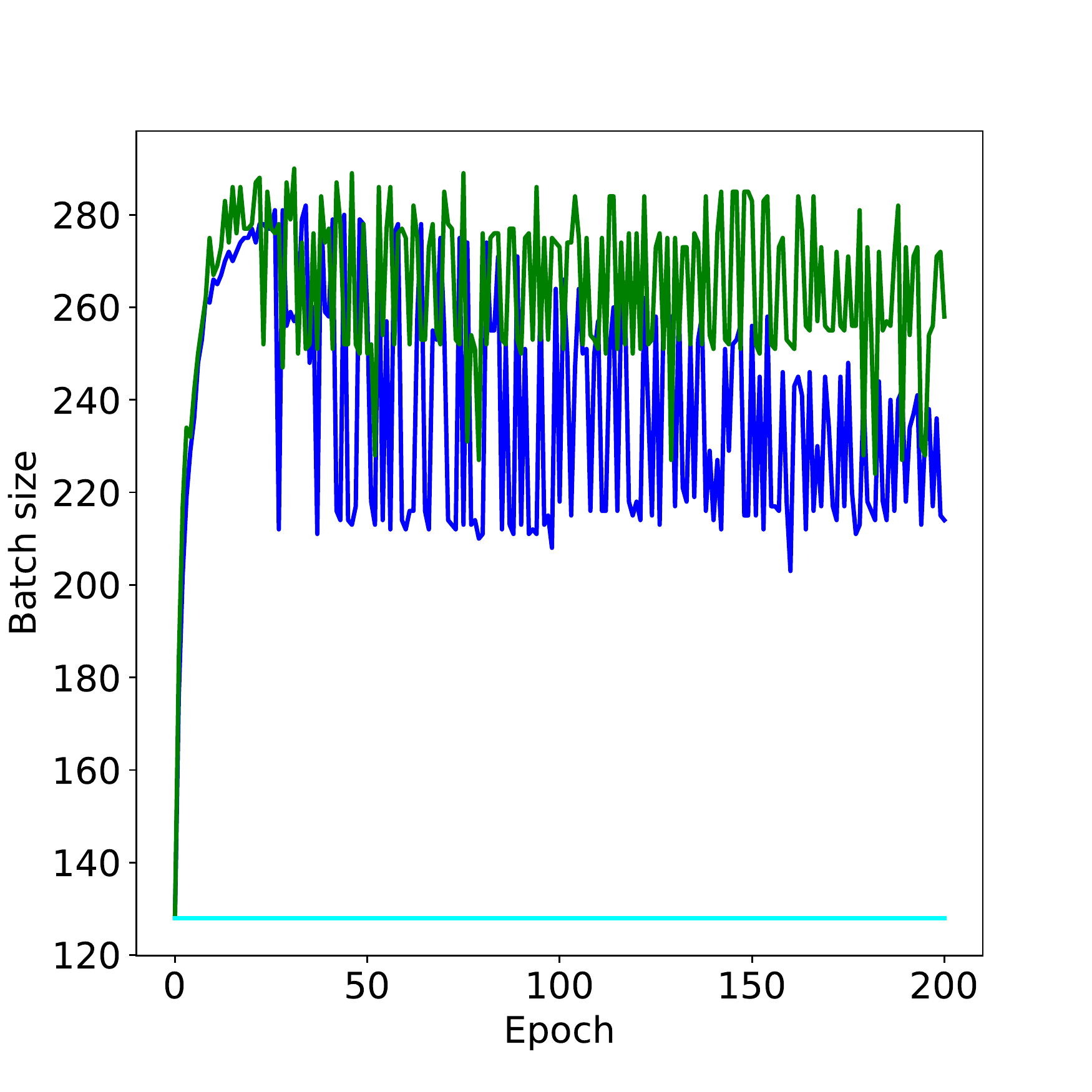} 
    \end{subfigure}
    %
     \begin{subfigure}[t]{0.234\textwidth}
        \includegraphics[width=\textwidth]{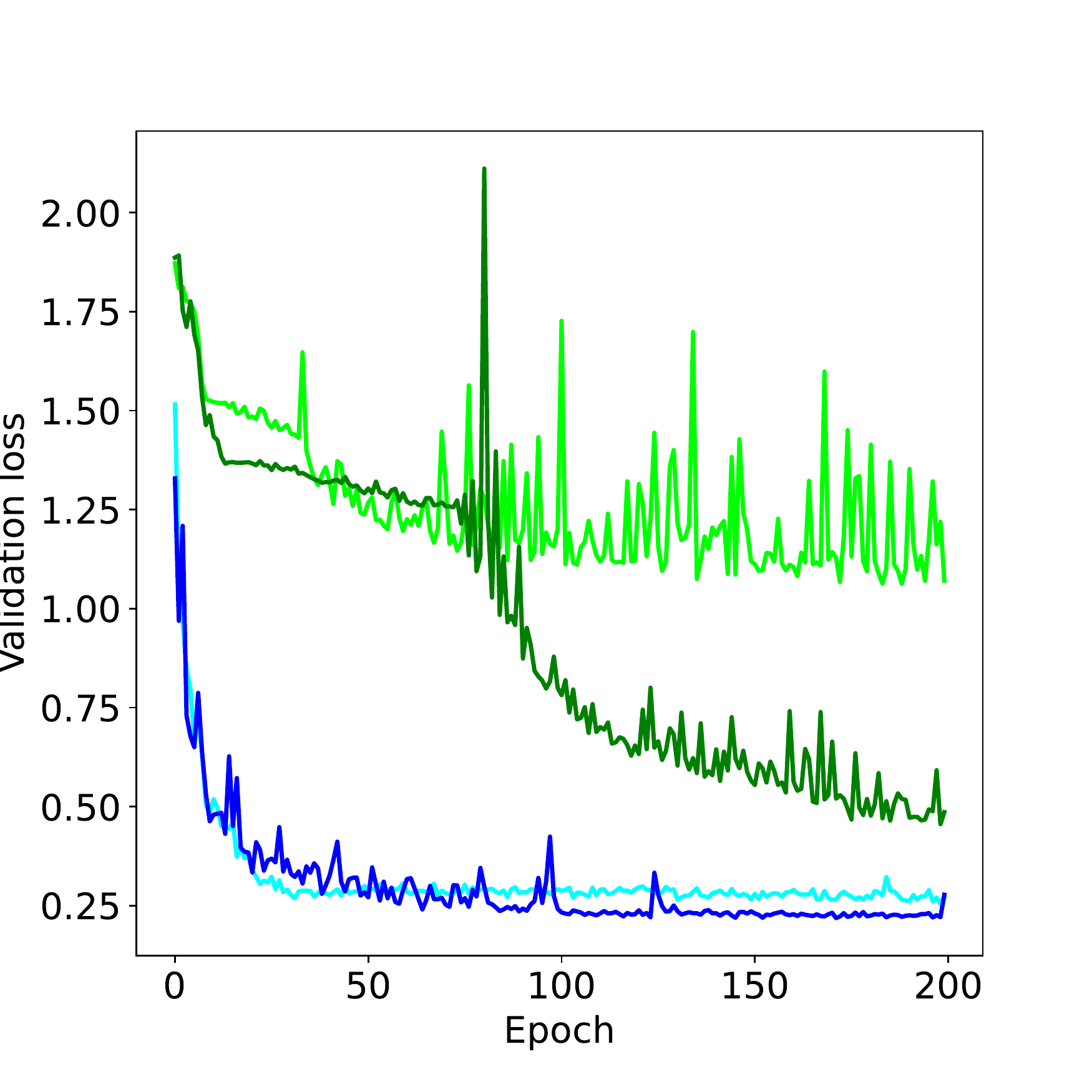} 
    \end{subfigure}
    \begin{subfigure}[t]{0.234\textwidth}
        \includegraphics[width=\textwidth]{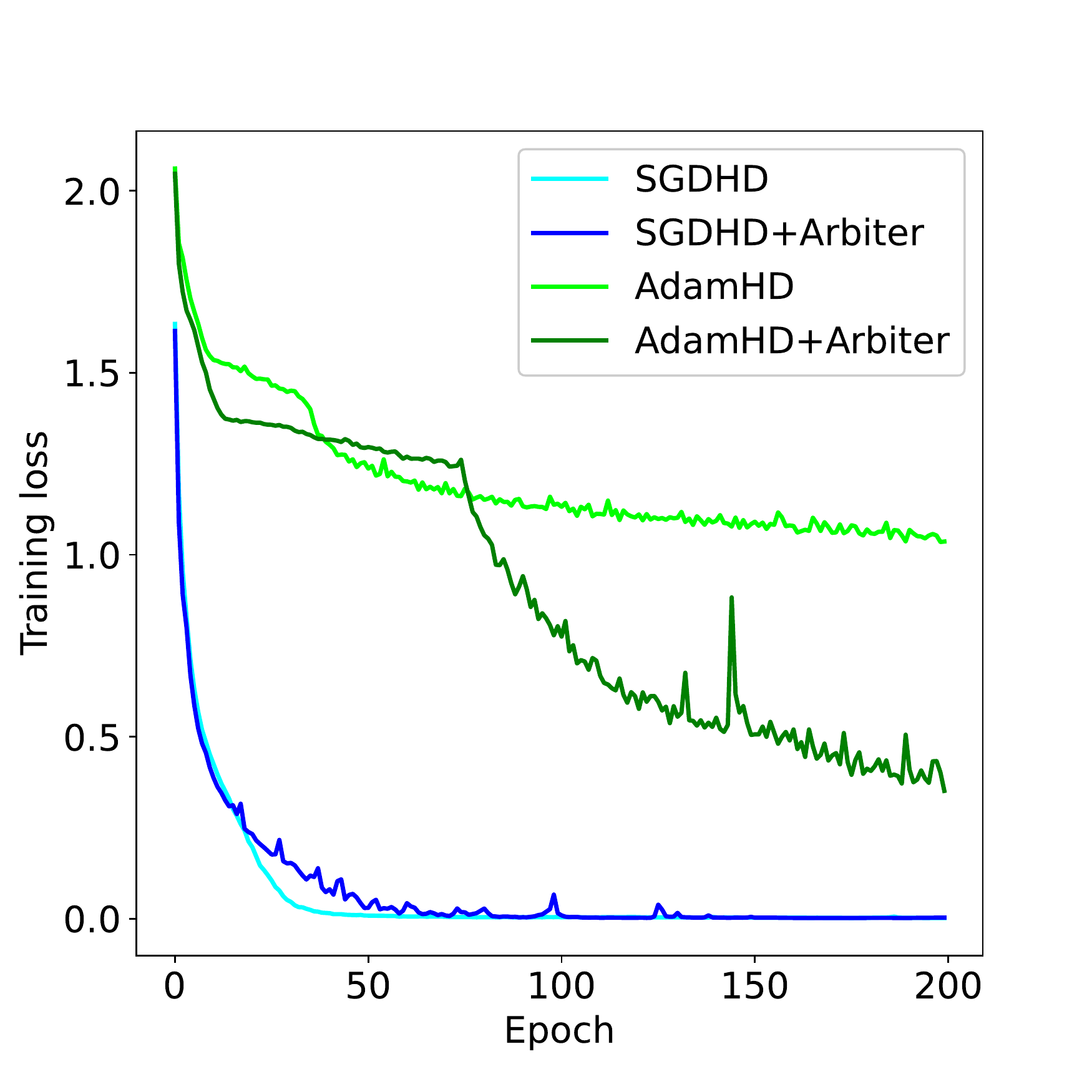} 
    \end{subfigure}
\end{center}
\vspace{-2.5mm}
\caption{\textbf{Controlling learning rate decay in hypergradient descent,} for the experiments in Sect. \textbf{\ref{sec:exp-lr}}. 
For both momentum and Adam, we observe that Arbiter has partial control over learning rate decay by adjusting gradient noise. Arbiter's choices in batch size adaptations lead to overall improved validation error in both optimizers.
The \textcolor{red}{red line} indicates the point in training when Arbiter begins to adapt the batch size.
}
\label{fig:exp-hd}
\end{figure}
\\\\
We experimented with hypergradient descent (HD) \cite{hd} using a large hyper-learning rate of $\beta=1e-4$ to decay the initial learning rate from $\eta=0.1$. This setup was chosen to promote the effects of short-horizon bias by emphasising the hypergradient. We implemented Arbiter in conjunction with both momentum (SGDHD) and Adam (AdamHD) HD optimization, which enabled us to study the impact of Arbiter's induced variance reduction. Experiments were run on CIFAR-10 with the Wide ResNet-16-4 architecture using an initial batch size of 128. In Figure \textbf{\ref{fig:exp-hd}}, we observe that Arbiter's control over the batch size dampens rapid learning rate decay, leading to improved generalization for both momentum and Adam. It is also interesting to note that Arbiter induces large fluctuations to the batch size later in training, which has the effect of speeding up learning rate decay. 

\section{Conclusions}
\label{sec:conc}
We presented Arbiter as a novel hyperparameter optimization algorithm for performing batch size scheduling in deep neural networks.
Uniquely, Arbiter learns scheduling heuristics by accessing rich gradient information within objective functions, made possible by a new learning process for HO called \textit{hyper-learning}, which avoids unrolled optimization by instead leveraging information learned from meta-objective responses.
Arbiter's promising results as a stand-alone batch size scheduler, and complementary tool to fixed scheduling and hypergradient descent of the learning rate, motivates two promising research directions.
First, to improve its learned heuristics with greater knowledge of the local geometry.
Second, to explore wider applications of hyper-learning, including discrete, continuous, and multi-level hyperparameter optimization.

\bibliographystyle{unsrt}
\bibliography{bib}

\end{document}